\documentclass[lettersize,journal]{IEEEtran}
\usepackage{amsmath,amsfonts}
\usepackage{algorithmic}
\usepackage{algorithm}
\usepackage{array}
\usepackage{textcomp}
\usepackage{stfloats}
\usepackage{url}
\usepackage{verbatim}
\usepackage{graphicx}
\usepackage{cite}
\usepackage{bm}
\usepackage{amssymb}
\usepackage{threeparttable}
\usepackage{float}
\usepackage{booktabs}
\usepackage{subfigure}
\usepackage{multirow}

\begin{document}

\title{Multi-level Cross-modal Feature Alignment via Contrastive Learning towards Zero-shot Classification of Remote Sensing Image Scenes }

\author{Chun Liu, Suqiang Ma, Zheng Li, Wei Yang and Zhigang Han  
\thanks{This paper was produced by the IEEE Publication Technology Group. They are in Piscataway, NJ.}
\thanks{Manuscript received xxx, 2022; revised xxx xx, 2022.}
\thanks{Chun Liu, Suqiang Ma, Zheng Li and Wei Yang are with School of Computer and Information Engineering, Henan Key Laboratory of Big Data Analysis and Processing, Henan Engineering Laboratory of Spatial Information Processing and Henan Industrial Technology Academy of Spatio-Temporal Big Data, Henan University, Zhengzhou 450046, China (e-mail: liuchun@henu.edu.cn; masuqiang@henu.edu.cn; lizheng@henu.edu.cn; weiyang@henu.edu.cn).}
\thanks{Zhigang Han is with College of Geography and Environmental Science and Henan Industrial Technology Academy of Spatio-Temporal Big Data, Henan University, Zhengzhou 450046, China (e-mail: zghan@henu.edu.cn);}
}

\markboth{Journal of \LaTeX\ Class Files,~Vol.~xx, No.~xx, xxx ~202x}%
{Shell \MakeLowercase{\textit{et al.}}: A Sample Article Using IEEEtran.cls for IEEE Journals}

\maketitle

\begin{abstract}
Zero-shot classification of image scenes which can recognize the image scenes that are not seen in the training stage holds great promise of lowering the dependence on large numbers of labeled samples. To address the zero-shot image scene classification, the cross-modal feature alignment methods have been proposed in recent years. These methods mainly focus on matching the visual features of each image scene with their corresponding semantic descriptors in the latent space. Less attention has been paid to the contrastive relationships between different image scenes and different semantic descriptors. In light of the challenge of large intra-class difference and inter-class similarity among image scenes and the potential noisy samples, these methods are susceptible to the influence of the instances which are far from these of the same classes and close to these of other classes. In this work, we propose a multi-level cross-modal feature alignment method via contrastive learning  for zero-shot classification of remote sensing image scenes. While promoting the single-instance level positive alignment between each image scene with their corresponding semantic descriptors, the proposed method takes the cross-instance contrastive relationships into consideration, and learns to keep the visual and semantic features of different classes in the latent space apart from each other.  Extensive experiments have been done to evaluate the performance of the proposed method. The results show that our proposed method outperforms state of the art methods for zero-shot remote sensing image scene classification. All the code and data are available at github https://github.com/masuqiang/MCFA-Pytorch.

\end{abstract}


\section{Introduction}
\IEEEPARstart{W}{ith} the rapid development of remote sensing observation technology,  remote sensing images have been greatly improved in spatial resolution. The images with high spatial resolution express more detailed features of the ground objects and their surrounding environments. Different ground objects with different spatial distribution relationships will form different high-level semantic scenes, which help people cross the gap between the underlying image features (e.g., color and texture) and the high-level semantics \cite{1}. Therefore, as an important manner to mine the higher-level scene semantics, image scene classification  is of great significance for remote sensing image understanding, and plays an important role in the fields of remote sensing image retrieval, disaster monitoring, and urban planning \cite{2}.

Because it is difficult to prepare sufficient samples for each class of image scenes for training, zero-shot learning (ZSL) methods which can recognize the image scenes that are not seen in the training stage, have extensive prospects for image scene classification\cite{4}. ZSL  aims to build a mapping between the visual space and the semantic space based on a large number of samples from seen classes, and transfers the model from seen classes to unseen classes in a data-free manner. In light of the difficulity to know whether the samples to be classified is seen or unseen in the real world, it is more applicable for the learned model to have the ability of classifying both seen and unseen classes. This more challenging problem is often called the generalized zero-shot learning (GZSL).

 To address the zero-shot image scene classification,   cross-modal feature alignment methods \cite{5,6} have been proposed in recent years. These methods learn to match the image scenes with their semantic descriptors in the latent space. From the perspective of each image scene, they mainly focus  on reducing the distance from the image scene to its corresponding semantic descriptor in the latent space. While, less attention has been paid to the contrastive relationships between different image scenes and different semantic descriptors, e.g., the distance from one image scene to other image scenes and from one semantic descriptor to other semantic descriptors.  However, due to the specific characteristics of remote sensing image scenes, there often exist large intra-class difference and inter-class similarity among image scenes \cite{6}.   In addition, there may be also noisy image  scenes. All these may affect the learning of the cross-modal alignment between the image scenes with their semantic descriptors in the latent space.

 In this work, we propose a multi-level cross-modal feature alignment method, named MCFA,  via contrastive learning \cite{7} for zero-shot classification of remote sensing image scenes. While promoting the alignment between each image scene with their corresponding semantic descriptor, we also take the cross-instance contrastive relationships into consideration to further improve the cross-modal feature alignment performance. Following the aligned VAEs method \cite{8}, the proposed method  integrates the single-instance level positive alignment losses and the cross-instance level contrastive alignment losses to constrain the model learning process.  While learning to algin the visual features of image scenes with their corresponding semantic features, the model can also learn to keep the visual and semantic features of different classes apart from each other in the latent space.

The remainder of this paper is structured as follows:  Section 2 details the proposed method; Section 3  presents our experiments and the results; and Section 4 are the conclusions of our work.

\section{Methodology}
In this section, we detail  the proposed method  by introducing the model framework and the losses used.

\begin{figure*}
\vspace{-0.3cm}
	\centering
	  \includegraphics[width=.9\textwidth,height=3in]{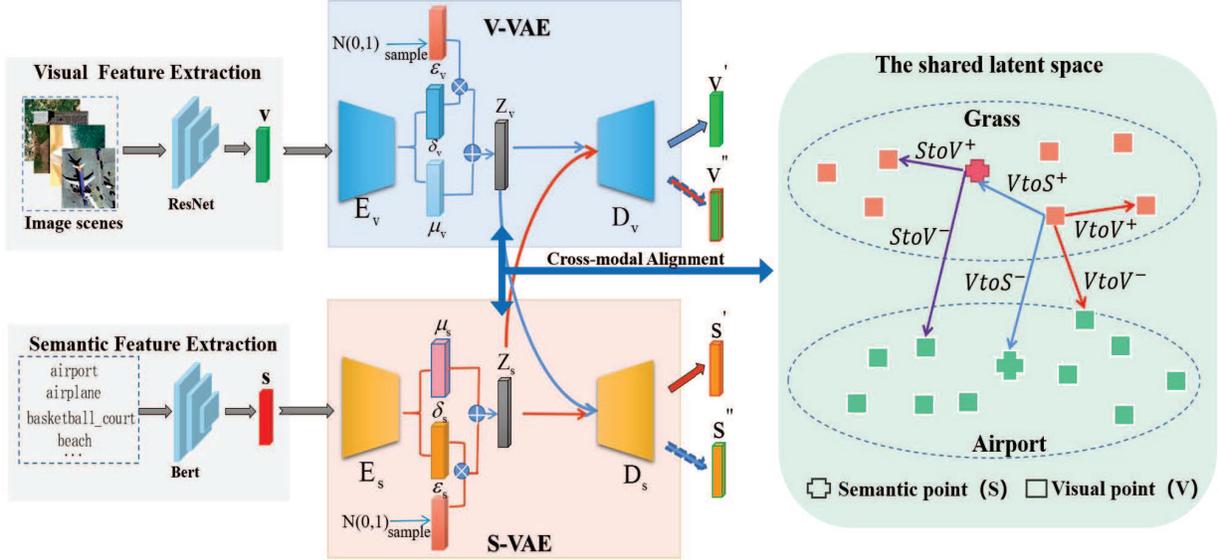}
	\caption{The overview of the proposed method. }
	\label{FIG:2}

\end{figure*}


\subsection{The Multi-level Cross-modal Feature Alignment Framework}

\textbf{Overview:} As shown in Fig. 1, there are two VAEs corresponding to two modalities in our model.  Each VAE consists of an encoder and a decoder. The inputs are the visual features of image scenes and the semantic features of scene classes, which are extracted respectively by the models such as ResNet\cite{9} and Bert \cite{10} in advance. With the inputs, the encoders project the visual features and the semantic features into the latent space. Then, the decoders reconstruct the visual features and the semantic features from the projected latent features.  Using the training set, it is expected that the VAEs can learn to enforce the cross-modal alignment between the two kinds of latent features embedded by the encoders. Once trained, the model can be used to generate the latent semantic features for unseen classes from their text descriptors. Under the ZSL setting,  a classifier can be trained by using the generated latent semantic features of unseen classes. While for the GZSL setting, the classifier will be trained with the latent semantic features from unseen classes but also the latent visual features from seen classes. Then, for one image scene from the testing set, its class can be inferred by generating its corresponding latent feature with the encoder and subsequently inputting the latent visual features into the classifier.

\textbf{Two level cross-modal feature alignment constrains: }For each class of image scenes, there are many visual instances but one semantic descriptor projected into the latent space, as shown in the right part of Fig. 1. The visual instances of one class are supposed to center around the semantic descriptor of the class in the latent space. But there may be some noise instances which are father to their own semantic descriptors but closer to the semantic descriptors of other classes. Particularly, from the perspective of one image scene, there are four kinds of relationships between the visual instances(V) and the semantic descriptors(S):

\begin{itemize}

\item \bm{$VtoS^{+}$}:  The relationship between the visual instance with its corresponding semantic descriptor.


\item \bm{$VtoS^{-}$}: The relationship between the visual instance with the semantic descriptors of other classes.


\item \bm{$VtoV^{+}$}: The relationship between the visual instance with other instances from the same class.


\item \bm{$VtoV^{-}$}: The relationship between the instance with the instances from other classes.


 \end{itemize}

Among these relationships, it is expected that  two items associated by the relationships of $VtoS^{+}$ and $VtoV^{+}$  are close as much as possible in the latent space. On the contrary, it is better to keep the items in relationships of $VtoS^{-}$ and $VtoV^{-}$  apart from each other as much as possible,
reducing the possibility of mis-matching between the items in the relationships of $VtoS^{+}$ and $VtoV^{+}$.

Moreover, from the perspective of one semantic descriptor, there are two more kinds of relationships(since each class has only one semantic descriptor, the class-to-class relationship is omitted.):
\begin{itemize}
\item \bm{$StoV^{+}$}: The relationship between the semantic descriptor and the visual instances from the same class. Obviously, this kind of relationships is equal to that of $VtoS^{+}$.


\item \bm{$StoV^{-}$}: The relationship between the semantic descriptor and the visual instances from other classes. This is contrastive to $StoV^{+}$, i.e., it is better to keep the two items associated by the relationship apart from each other.
 \end{itemize}

In current works like \cite{6}, much attention has been paid to the relationship of $VtoS^{+}$. They focus on positively promoting the single-instance level cross-modal feature alignment, i.e., each visual instance matches with its semantic descriptor. In this paper, we propose to take all these relationships into consideration. It is obvious that these relationships form three contrastive relationships, i.e., $VtoS^+$ vs. $VtoS^-$, $VtoV^+$ vs. $VtoV^-$, and $StoV^+$ vs. $StoV^-$,  which cross multi instances or their corresponding class descriptors. With these contrastive relationships as new constrains, we can deal with the challenge of large intra-class difference and inter-class similarity among the remote sensing image scenes. It will constrain the influence of noisy samples or outliers on the learning of cross-modal feature alignment. Therefore, we propose the  \textbf{cross-instance level contrastive alignment losses } to complement the  \textbf{single-instance level positive alignment losses}. Next, we detail these two levels of loss constrains.

\subsection{Single-instance Level Positive Alignment Losses}
The single-instance level loss constrains promote the alignment between each image scene and its semantic descriptor in the latent space. Following the aligned VAEs method \cite{8}, there are three kinds of losses for this purpose: the VAE loss, the cross-modal feature reconstruction loss and  the cross-modal distribution alignment loss. The definitions of them are described as follows.

\subsubsection{The VAE Loss}
For each VAE shown in Fig. 1, the encoder maps the input features into the latent space, and the decoder strives to reconstruct them from the mapped latent features. The VAE loss measures the reconstruction quality, and also expects that the data distribution in the latent space obeys the standard Gaussian distribution, which  is defined as Eq.(1).

\begin{equation}
\begin{aligned}
\ell_{VAE} &= E_{p_{E_v} (z_v|v)}[log p_{D_v} (v|z_v)] + E_{p_{E_s}(z_s|s)}[log p_{D_s}(s|z_s)] - \\
&D_{KL}(p_{E_v}(z_v|v)||p(z_v)) - D_{KL}(p_{E_s}(z_s|s)||p(z_s))
\end{aligned}
\end{equation}

 $E_{p_{E_v} (z_v|v)}[log p_{D_v} (v|z_v)]$ and $E_{p_{E_s}(z_s|s)}[log p_{D_s}(s|z_s)]$ are the reconstruction losses, measuring the difference between the input data and the reconstructed data. While, the items of $D_{KL}(p_{E_v}(z_v|v)||p(z_v))$ and $D_{KL}(p_{E_s}(z_s|s)||p(z_s))$ are the Kullback-Leibler(KL) divergence losses, referring to the difference between the distribution $p_E(z|x)$ of the generated latent variable $z$ in the latent space and the unit Gaussian distribution $p(z)$.

\subsubsection{The Cross-modal Feature Reconstruction Loss}

 The cross-modal feature reconstruction loss shown in Eq.(2) aims to constrain the latent features to align with each other in the latent space.  $N$  denotes the number of training instances in one batch, and $v^i$ and $s^i$ represent the visual feature and semantic feature of $i^{th}$ instance.

\begin{equation}
\begin{aligned}
\ell_{CMFR} =\sum_{i=1}^N|v^i-D_v(E_s(s^i))|+|s^i-D_s(E_v(v^i))|
\end{aligned}
\end{equation}

\subsubsection{The Cross-modal Distribution Alignment Loss}
The cross-modal distribution alignment loss is to constrain the distribution alignment between each image scene and its  semantic descriptor. Its definition is shown in Eq.(3) where $\mu^{i}$ and $\sqrt{E^{i}}$ represent the the mean and standard deviation of the feature distribution in the latent space corresponding to $i^{th}$ image scene.

\begin{equation}
\begin{aligned}
\ell_{CMDA} =\sum_{i=1}^N\sqrt{||\mu_{v}^{i}-\mu_{s}^{i}||_{F}^{2}+||\sqrt{E_{v}^{i}}-\sqrt{E_{s}^{i}}||_{F}^{2}}
\end{aligned}
\end{equation}

\subsection{Cross-instance Level Contrastive Alignment Losses}
The cross-instance level loss constrains  aim to promote the positive feature matching, i.e., $VtoS^{+}$, $VtoV^{+}$ and $StoV^{+}$, and weaken the negative feature matching, i.e., $VtoS^{-}$, $VtoV^{-}$ and $StoV^{-}$. Corresponding to these three kinds of contrastive relationships, three cross-instance level loss constrains are introduced based on the supervised contrastive learning \cite{7}: visual-to-visual contrastive loss, visual-to-semantic contrastive loss, and semantic-to-visual contrastive loss. They are defined as follows.

\subsubsection{The Visual-to-visual Contrastive Loss}
Taking an image scene as anchor, the visual-to-visual contrastive loss tends to make it close to the image scenes from the same class in the latent space, and apart from the image scenes from other classes. Given a batch of image scenes whose normalized latent visual features are $z^{v}$, the visual-to-visual contrastive loss is defined by Eq.(4).

\begin{equation}
 \ell_{VtoV}=-\sum_{i\in I}\frac{1}{|P(i)|}\sum_{p\in P(i)} log\frac{exp(z^{v}_{i} \bullet z^{v}_{p}/\tau)}{\sum_{a\in I \backslash i}exp(z^v_{i}\bullet z^v_a/\tau)}
\end{equation}

In above definition, $i\in I=|z^v|$ is the index of an arbitrary visual instances in the batch. $P(i)=\{p\in I\backslash i|y_p = y_i\}$ is the set of indices excluding $i$, where the instances indexed by the items in the set are from the same class as that of the $i^{th}$ instance. And, $\tau \in R^+$ is a temperature parameter controlling the concentration of contrastive losses on hard samples.

\subsubsection{The Visual-to-semantic Contrastive Loss}
Also taking an image scene as anchor, the visual-to-semantic contrastive loss aims to promote its alignment with its corresponding semantic descriptor, and weaken its alignment with the semantic descriptors of other classes in the latent space. Given a batch of image scenes whose normalized latent visual features are $z^{v}$, and normalized corresponding semantic features are $z^s$, the  visual-to-semantic contrastive loss is defined by Eq.(5).

\begin{equation}
 \ell_{VtoS}=-\sum_{i\in I}log\frac{exp(z^{v}_{i} \bullet z^{s}_{s(i)}/\tau)}{\sum_{k=1}^{|z^s|}exp(z^v_{i}\bullet z^s_k/\tau)}
\end{equation}

Where $s(i)$ means the index of the corresponding semantic descriptor of the $i^{th}$ image scene.

\subsubsection{The Semantic-to-visual Contrastive Loss}
The semantic-to-visual contrastive loss shifts focus to the alignment from the semantic descriptors to the image scenes. It is to promote the alignment of each semantic descriptor with the image scenes from the same class, and weaken the alignment with the image scenes of other classes in the latent space. Its definition is shown in Eq.(6).

\begin{equation}
 \ell_{StoV}=-\sum_{i\in J}\frac{1}{|P(i)|}\sum_{p\in P(i)} log\frac{exp(z^{s}_{i} \bullet z^{v}_{p}/\tau)}{\sum_{a\in I}exp(z^s_{i}\bullet z^v_a/\tau)}
\end{equation}

Where $i\in J=|z^s|$ is the index of an arbitrary semantic feature in the batch.   $P(i)=\{p\in I |y_p = i\}$ is the set of indices, where the features indexed by the items in the set are from the classes of the $i^{th}$ semantic descriptor.

\subsection{The Overall Loss}

With all above defined loss functions, the overall loss function of our model is defined as Eq.(7) where $\lambda_i(i=1,2,3,4,5)$ is the loss weight factor.

 \begin{equation}
\begin{aligned}
 \ell_{total}=\ell_{VAE}+{\lambda_1}\ell_{CMFR}+{\lambda_2}\ell_{CMDA}+{\lambda_3}\ell_{VtoV} \\
 +{\lambda_4}\ell_{VtoS}+{\lambda_5}\ell_{StoV}
\end{aligned}
\end{equation}

\section{Experiments}

To validate the proposed method, several experiments have been done in our work. In this section, we describe the experimental setup and the experimental results.

\subsection{Experimental Setup}

\subsubsection*{Dataset and Preprocessing}

 We take the dataset which has been used in the work of Li, et al. \cite{6} for our experiments.  There are 70 classes in the dataset, and 800 image scenes per class with the size of 256 $\times$ 256 pixels.

As can be seen in Fig. 1, instead of directly inputting the image scenes into our model, we first use an extractor to extract their visual features. In our work, we use the classical model of ResNet18 \cite{9} for this purpose, which has been pretrained on the ImageNet dataset and can be  accessed in the Python environment. The extracted visual features are with 512 dimensions. Similarly, the semantic features of scene classes also need to be extracted in advance.  We adopt the sematic features which are extracted by Bert \cite{10} and have been used by the work of Li et al. \cite{6} for experiments. They are 1024 dimensional features extracted from a set of sentences describing the image scene classes.

\subsubsection*{Implementation and Configuration}
The model we used in our work is comprised by two VAEs and a softmax classifier. The two VAEs are for two different modalities, i.e., the visual and the semantic, respectively. All the encoders and decoders in both VAEs are the neural networks with only one hidden layer. For visual modality,  the hidden layers of the encoder and decoder are  implemented with 512 dimension. Meanwhile, the hidden layers of the encoder and decoder for the semantic modality are implemented with 256 dimension.  When using the generated latent features to train a classifier for predicting the classes of testing image scenes, the softmax classifier is applied. There is a set of parameters in our proposed method. The settings of them are shown in Table I. And 50 epochs are taken to train the model.

\begin{table}[h]
  \caption{The settings of key parameters}
  \centering
  \setlength{\tabcolsep}{0.9mm}{
  \begin{tabular}{llll}
   \toprule
    \multicolumn{1}{l}{\multirow{2}{*}{Parameters}} & \multicolumn{1}{l}{\multirow{2}{*}{Description}}  & \multicolumn{2}{c}{Settings}\\
    \cmidrule(r){3-4}
    & & ZSL & GZSL \\
    \midrule
    $\lambda_1$  &The weight factor to $\ell_{CMFR}$ & 10 & 1\\
    $\lambda_2$ & The weight factor to $\ell_{VSDM}$ & 1 & 1\\
    $\lambda_3$ & The weight factor to $\ell_{VtoV}$ & 100 & 0.1\\
    $\lambda_4$ & The weight factor to $\ell_{VtoS}$ & 100 & 1\\
    $\lambda_5$ & The weight factor to $\ell_{StoV}$ & 10 & 1\\
    $\tau$ & The temperature parameter in Eq.(4) & 2 & 2 \\
    $c$ & Number of classes in batch & 5 & 5 \\
    $k$ & Number of instances per class in batch &5 & 10 \\
    $\gamma$ & The size of latent space & 64 & 64\\
    \bottomrule
  \end{tabular}
}
\end{table}

We have divided the dataset according to the ratios of 60/10, 50/20 and 40/30 to obtain the seen and unseen classes of image scenes. For each ratio, the average of the classification accuracies over five seen/unseen splits is taken as the final result. And under the GZSL setting, we select 600 image scenes of each seen class for training, and the remaining 200 image scenes are used for testing. We report the overall accuracy for ZSL setting, and the harmonic mean accuracy for GZSL setting.

\subsection{Comparison with Related Methods}
To validate the performance of our method, we have taken several classical zero-shot or generalized zero-shot methods as baselines for comparison.   The results of the comparisons are shown in Table II. It can be seen that for the zero-shot and generalized zero-shot classification of remote sensing image scenes, our proposed method outperforms these baseline methods.

\begin{table}[ht]
	\caption{The results of the comparison with related methods. \label{tab:table1}}
	\centering
\setlength{\tabcolsep}{0.6mm}{
	\begin{tabular}{llll|lll}
		\toprule
        \multicolumn{1}{l}{\multirow{2}{*}{Methods}} & \multicolumn{3}{c}{ZSL} & \multicolumn{3}{c}{GZSL}\\
        \cmidrule(r){2-4}
         \cmidrule(r){5-7}
		 &  60/10 & 50/20 & 40/30 &  60/10 & 50/20 & 40/30 \\
		\hline
		SAE[11]  &0.1362     & 0.097      &0.0503  &0.0157 & 0.0189 &0.0017  \\
		
		GDAN[12]  &0.3249    & 0.1753      &0.1137 &0.1038  & 0.1151  &0.0867\\
		
		CADA-VAE[8]  &0.1721   & 0.0943    &0.0543  &0.1315  & 0.0953  &0.0542 \\
	
		DAN[6] &0.2601     & 0.1574         &0.1115  &0.1736  & 0.1274  &0.1072\\
        CE-GZSL[13]  &0.3002    & 0.1541    &0.1213  &0.2297 & 0.1499   &0.1107 \\
        \hline
		MCFA    &0.3661      & 0.2063     &0.1385  &0.2352    & 0.1632  &0.1274\\
		\bottomrule
	\end{tabular}
}
\end{table}

Particularly, when compared with  CE-GZSL \cite{13} which has also introduced the contrastive learning into GZSL task, our method has shown better performance. CE-GZSL  extended GAN model by selecting  several samples from different classes as the contrastive samples and several class descriptors of different classes as the contrastive descriptors for each instance. Different from that,  we have followed the way of cross-modal feature alignment to generate intermediate latent features for unseen classes and use them to train the classifier, instead of generating visual features for unseen classes directly from their semantic descriptors. The generated intermediate latent features may contain the important visual information but also the semantic information. Moreover, more contrastive relationships under the cross-modal feature alignment setting are also taken into account. These may lead to that our proposed method resolves the zero-shot image scene classification better.

\subsection{Ablation Experiments}

As aforementioned, three new  contrastive alignment loss functiones have been introduced to constrain the learning process in our method. They are  the visual-to-visual contrastive loss $\ell_{VtoV}$, the visual-to-semantic contrastive loss $\ell_{VtoS}$, and the semantic-to-visual contrastive loss $\ell_{StoV}$.  To investigate the effectiveness of these components to the performance of image scene classification, the ablation experiment has been done. Table III shows the results under the seen/unseen ratio of 60/10.

\begin{table}[ht]
	\caption{The results of ablation experiment\label{tab:table1}}
	\centering
	\setlength{\tabcolsep}{0.8mm}{
		\begin{tabular}{lllllll|ll}
			\toprule
		  Variants	&$\ell_{VAE}$ &$\ell_{CMFR}$ &$\ell_{CMDA}$ & $\ell_{VtoV}$ &$\ell_{VtoS}$ &$\ell_{StoV}$ & ZSL &GZSL \\
			\hline
			$v_0$ &\checkmark & \checkmark & \checkmark &  & &  &0.3134   & 0.2243    \\
			$v_1$  &\checkmark & \checkmark & \checkmark & \checkmark  &   &   &0.336  & 0.2319    \\
			$v_2$ &\checkmark & \checkmark & \checkmark  &  &\checkmark  &   &0.3439  & 0.2305     \\
			$v_3$ &\checkmark & \checkmark & \checkmark &  &  & \checkmark  & 0.3327  & 0.2321  \\
			\hline

			$v_4$ &\checkmark & \checkmark & \checkmark  & \checkmark  & \checkmark  &  &0.3558     & 0.2328   \\
			$v_5$(ours) &\checkmark & \checkmark & \checkmark & \checkmark  & \checkmark  & \checkmark   & 0.3661  & 0.2352   \\
			\bottomrule
		\end{tabular}
	}
\end{table}

When comparing the variants of $v_1$, $v_2$ and $v_3$ to that of $v_0$, the results show that all these contrastive losses are useful for the zero-shot classification of remote sensing image scenes. As shown in Table III, the performance has reached from $31.34\%$ to over $33\%$ under ZSL setting, and from $22.43\%$ to over $23\%$ under the GZSL setting when applying these contrastive losses. When comparing these contrastive losses, it can be seen that the visual-to-semantic contrastive loss $\ell_{VtoS}$  produces more contribution  under the ZSL setting, but under GZSL setting,  the semantic-to-visual contrastive loss $\ell_{StoV}$ plays a more important role.  This indicates that there are more benefits from the contrastive losses considering the cross-modal relationships between the image scenes and their semantic descriptors.

Based on the variant of $v_1$, variant $v_4$ further includes the visual-to-semantic contrastive loss $\ell_{VtoS}$, and variant $v_5$ takes all the three contrastive losses simultaneously. The results show that the classification performance experiences improvement when using these contrastive losses together. For example, when applying the contrastive losses $\ell_{VtoV}$ and $\ell_{VtoS}$ together, the performance rises from $33.6\%$ to $35.58\%$ under ZSL setting, and from $23.19\%$ to $23.28\%$ under GZSL setting. When applying all three contrastive losses, the performance further improves to $36.61\%$ under  ZSL setting and $23.52\%$ under GZSL setting.  This demonstrates that these losses do not conflict with each other and can be collaboratively utilized  to obtain better zero-shot classification performance of remote sensing image scenes.

\section{Conclusion}

This paper proposes a multi-level cross-modal feature alignment method via contrastive learning for zero-shot classification of remote sensing image scenes. It uses two VAEs to project the visual features of image scenes and the semantic features of scene classes into a shared latent space, and learns to achieve the alignment between them in that space. Taking the cross-instance contrastive relationships into consideration, it proposes cross-instance level contrastive alignment losses to complement the single-instance level positive alignment losses. While learning to algin the visual features of image scenes with their corresponding semantic features, it can also learn to keep the visual and semantic features of different classes in the latent space apart from each other. It shows the promise to address the challenge of large intra-class difference and inter-class similarity among image scenes and the potential noisy samples. Based on the widely used dataset, extensive experiments have been done to evaluate the performance of the proposed method. The results show that our proposed method outperforms state of the art methods for the zero-shot classification of remote sensing image scenes.

%
\bibliographystyle{IEEEtran}


\begin{thebibliography}{}
\providecommand{\url}[1]{#1}
\csname url@samestyle\endcsname
\providecommand{\newblock}{\relax}
\providecommand{\bibinfo}[2]{#2}
\providecommand{\BIBentrySTDinterwordspacing}{\spaceskip=0pt\relax}
\providecommand{\BIBentryALTinterwordstretchfactor}{4}
\providecommand{\BIBentryALTinterwordspacing}{\spaceskip=\fontdimen2\font plus
\BIBentryALTinterwordstretchfactor\fontdimen3\font minus
  \fontdimen4\font\relax}
\providecommand{\BIBforeignlanguage}[2]{{%
\expandafter\ifx\csname l@#1\endcsname\relax
\typeout{** WARNING: IEEEtran.bst: No hyphenation pattern has been}%
\typeout{** loaded for the language `#1'. Using the pattern for}%
\typeout{** the default language instead.}%
\else
\language=\csname l@#1\endcsname
\fi
#2}}
\providecommand{\BIBdecl}{\relax}
\BIBdecl

\end{thebibliography}


\begin{thebibliography}{99}


\bibitem{1}
D. Bratasanu,  I. Nedelcu  and  M. Datcu, ``Bridging the semantic gap for satellite image annotation and automatic mapping applications," \textit{ IEEE Journal of Selected Topics in Applied Earth Observations and Remote Sensing}, vol. 4, no. 1, pp. 193--204, 2011.

\bibitem{2}
G. Cheng, L. Guo,  T. Zhao, J. Han, H. Li and J. Fang,  ``Automatic landslide detection from remote-sensing imagery using a scene classification method based on BoVW and pLSA," \textit{ International Journal of Remote Sensing}, vol. 34,  pp. 45--59, 2013.

\bibitem{4}
A. Li, Z. Lu, L. Wang, T. Xiang and J. Wen, ``Zero-shot scene classification for high spatial resolution remote sensing images," \textit{ IEEE Transactions on Geoscience and Remote Sensing}, vol. 55, no. 7, pp. 4157--4167, 2017.

\bibitem{5}
Y. Li, Z. Zhu, J. Yu and Y. Zhang, ``Learning deep cross-modal embedding networks for zero-shot remote sensing image scene classification," \textit{ IEEE Transactions on Geoscience and Remote Sensing}, vol. 59, no. 12, pp. 10590--10603, 2021.

\bibitem{6}
Y. Li, D. Kong, Y. Zhang,  Y. Tan and  L. Chen,  ``Robust deep alignment network with remote sensing knowledge graph for zero-shot and generalized zero-shot remote sensing image scene classification," \textit{ ISPRS Journal of Photogrammetry and Remote Sensing}, vol. 179, pp. 145--158, 2021.

\bibitem{7}
P. Khosla, P. Teterwak, C. Wang, A. Sarna, Y. Tian, P. Isola, A. Maschinot, D. Krishnan and C. Liu, ``Supervised contrastive learning,"  arXiv:2004.11362, 2020.

\bibitem{8}
E. Schönfeld, S. Ebrahimi, S. Sinha, T. Darrell and Z. Akata, ``Generalized zero- and few-shot learning via aligned variational autoencoders," in \textit{IEEE/CVF Conference on Computer Vision and Pattern Recognition}, pp. 8239--8247, 2019.

\bibitem{9}
K. He, X. Zhang, S. Ren and J. Sun,  ``Deep residual learning for image recognition," in \textit{ Proceedings of the IEEE  Conference on Computer Vision and Pattern Recognition}, pp. 770--778, 2016.

\bibitem{10}
J. Devlin, M. Chang, K. Lee and K. Toutanova,  ``Bert: Pre-training of deep bidirectional transformers for language understanding,"  arXiv:1810.04805, 2018.

\bibitem{11}
E. Kodirov, T. Xiang and S. Gong, ``Semantic autoencoder for zero-shot learning," in \textit{Proceedings of the IEEE onference on Computer Vision and Pattern Recognition}, pp. 3174--3183, 2017.

\bibitem{12}
H. Huang, C. Wang, P. S. Yu and C. Wang, ``Generative dual adversarial network for generalized zero-shot learning," in \textit{ IEEE/CVF Conference on Computer Vision and Pattern Recognition}, pp. 801--810, 2019.


\bibitem{13}
Z. Han, Z. Fu, S. Chen and J. Yang, ``Contrastive embedding for generalized zero-shot learning," in \textit{IEEE/CVF Conference on Computer Vision and Pattern Recognition}, arXiv: 2103.16173,  2021.






\end{thebibliography}






\vfill

\end{document}